\documentclass{article}

\PassOptionsToPackage{numbers, compress}{natbib}
\usepackage[sglblindworkshop, final]{neurips_2025}
\usepackage[utf8]{inputenc} % allow utf-8 input
\usepackage[T1]{fontenc}    % use 8-bit T1 fonts
\usepackage{url}            % simple URL typesetting
\usepackage{booktabs}       % professional-quality tables
\usepackage{amsfonts}       % blackboard math symbols
\usepackage{nicefrac}       % compact symbols for 1/2, etc.
\usepackage{microtype}      % microtypography
\usepackage{xcolor}         % colors
\usepackage{amsmath}
\usepackage{amssymb}
\usepackage{bm}
\usepackage{graphicx}
\usepackage{multirow}
\usepackage{float}
\usepackage{color}

\title{On Evaluation of Unsupervised Feature Selection for Pattern Classification}
\workshoptitle{AI for Tabular Data workshop at EurIPS 2025}
\author{%
  Gyu-Il Kim\textsuperscript{1}, Dae-Won Kim\textsuperscript{2}, Jaesung Lee\textsuperscript{1}\thanks{Corresponding author.} \\
  \textsuperscript{1}Department of Artificial Intelligence, Chung-Ang University, Republic of Korea \\
  \textsuperscript{2}School of Computer Science and Engineering, Chung-Ang University, Republic of Korea\\
  \texttt{\{gyu6491, dwkim, curseor\}@cau.ac.kr} \\
}

\begin{document}

\maketitle

\begin{abstract}
Unsupervised feature selection aims to identify a compact subset of features that captures the intrinsic structure of data without supervised label. Most existing studies evaluate the performance of methods using the single-label dataset that can be instantiated by selecting a label from multi-label data while maintaining the original features. Because the chosen label can vary arbitrarily depending on the experimental setting, the superiority among compared methods can be changed with regard to which label happens to be selected. Thus, evaluating unsupervised feature selection methods based solely on single-label accuracy is unreasonable for assessing their true discriminative ability. This study revisits this evaluation paradigm by adopting a multi-label classification framework. Experiments on 21 multi-label datasets using several representative methods demonstrate that performance rankings differ markedly from those reported under single-label settings, suggesting the possibility of multi-label evaluation settings for fair and reliable comparison of unsupervised feature selection methods.
\end{abstract}

%%%%%%%%%%%%%%%%%%%%%%%%%%%%%%%%%%%%%%%%%%%%%%%%%%%%%%%%%%%%INTRODUCTION%%%%%%%%%%%%%%%%%%%%%%%%%%%%%%%%%%%%%%%%%%%%%%%%%%%%%%%%%%%%
\section{Introduction}
Unsupervised Feature Selection (UFS) aims to identify a compact yet informative subset of features that effectively represents the intrinsic structure of data without any label information. Because real-world datasets often include a large number of redundant or irrelevant features, UFS serves as a crucial preprocessing step for improving interpretability and subsequent learning performance \cite{li2012unsupervised}.

The prevailing evaluation paradigm in UFS research implicitly assumes the superiority among UFS methods can be evaluated under the single-label setting \cite{karami2023unsupervised}. In practice, real-world data are often multi-label in nature, where feature sets may correspond to multiple valid label combinations. In this regard, a single-label dataset can be viewed as an instantiation of selecting a label from a multi-label dataset based on some intention while maintaining the original feature set, where the discarded labels are unknown. Consequently, the performance reported under single-label evaluation may not reflect the true representational capability of the selected feature subset but may rather depend on the luck regarding the arbitrarily chosen label.

To address this overlooked issue, this study revisits the evaluation framework of UFS by adopting a multi-label evaluation paradigm. We investigate how existing UFS models perform when evaluated in a multi-label classification environment. By employing representative multi-label evaluation measures such as Hamming Loss, Ranking Loss, One-Error, and Multi-Label Accuracy, we systematically analyze whether the relative performance rankings of existing UFS methods remain consistent when the evaluation shifts from single-label to multi-label settings.

%%%%%%%%%%%%%%%%%%%%%%%%%%%%%%%%%%%%%%%%%%%%%%%%%%%%%%%%%%%%RELATED WORK%%%%%%%%%%%%%%%%%%%%%%%%%%%%%%%%%%%%%%%%%%%%%%%%%%%%%%%%%%%%
\section{Related Work}
Unsupervised FS has been widely studied as a preprocessing strategy for dimensionality reduction and data interpretation. Existing approaches are generally categorized into graph-based, information-theoretic, and evolutionary frameworks. Graph-based methods, such as Laplacian Score and Multi-Cluster Feature Selection (MCFS), preserve local manifold structures by constructing affinity graphs that capture neighborhood similarities, while methods like Unsupervised Discriminative Feature Selection (UDFS) introduce spectral regularization to enhance discriminative capacity \cite{mcfs,udfs}.

Information-theoretic approaches aim to maximize feature dependency or joint entropy to identify informative subsets, and evolutionary or memetic methods such as Robust Unsupervised Feature Selection
 (RUFS) and Nonnegative Discriminative Feature Selection (NDFS) employ stochastic search and iterative refinement for global optimization \cite{rufs,ndfs}. Our previous work on Pattern Discrimination Power (PDP)-based FS demonstrated that maximizing joint entropy enhances the intrinsic discriminability of data without supervision \cite{pdp}.

Despite these advances, the performance of most methods has been evaluated only under single-label settings, typically by combining selected features with simple classifiers such as k-Nearest Neighbor, Naive Bayes, or Decision Tree. This evaluation scheme assumes that each instance belongs to a single class, disregarding that real-world data often exhibit multi-label associations, where one instance can correspond to multiple categories simultaneously. Consequently, single-label measures such as accuracy or NMI cannot fully reflect the structural generalization or representational robustness of the selected features.

To the best of our knowledge, no prior study has systematically examined FS performance under a multi-label evaluation framework. This study fills this gap by analyzing representative methods using multi-label measures and reveals how the evaluation paradigm itself can influence the perceived superiority of existing FS methods.
%%%%%%%%%%%%%%%%%%%%%%%%%%%%%%%%%%%%%%%%%%%%%%%%%%%%%%%%%%%%METHODOLOGY%%%%%%%%%%%%%%%%%%%%%%%%%%%%%%%%%%%%%%%%%%%%%%%%%%%%%%%%%%%%

\section{Methodology}
This study redefines the evaluation framework of UFS to address its practical inconsistency under multi-label conditions. We analyze how representative UFS methods perform when their selected features are evaluated using a multi-label classification setting.

Traditional UFS evaluation is typically performed under a single-label classification setting. In the real world, an object $x$ can typically be assigned to multiple different labels $L = \{l_1, \ldots, l_{\vert L \vert}$. According to this nature, a multi-label dataset that consists of multiple features $F$ and labels $L$ can be created. Thus, multiple single-label datasets can be derived by including the original feature set $F$ as it is while selecting a label $l^* \in L$. Because $l^*$ will be chosen by some intention that may be unknown to the observer, this process can be a random process. This randomness may result in biased evaluation or misleading conclusions. For example, a feature subset appearing to perform well under one label might yield incompetent performance if another label is considered.

All UFS methods are trained in a fully unsupervised manner without using label information. For each dataset, a fixed number of top-k features is selected based on each method’s scoring criterion. The selected features are then evaluated through a multi-label classification model, such as the ML-kNN classifier, which predicts multiple label memberships simultaneously. Each classifier is trained and tested under conditions to ensure fair comparison across FS methods.

To quantify the performance of UFS under multi-label conditions, we adopt four representative measures widely used in multi-label learning. The definition of Hamming Loss is given as follows.
\begin{equation}
{hloss}(h)=\frac {1}{p}\sum _{i=1}^p \frac {1}{q}~|h({\boldsymbol x}_i)\Delta Y_i|.
\end{equation}
Hamming Loss measures the proportion of misclassified labels among all possible label assignments. A value of 0 indicates that every label of each instance is correctly predicted, while 1 represents complete disagreement between prediction and ground truth. A smaller Hamming Loss implies better multi-label classification performance. The definition of One-Error is given as follows.
\begin{equation}
{one-error}(f)=\frac {1}{p}\sum _{i=1}^p [\![\left [\arg \max \nolimits _{y\in \mathcal {Y}}f({\boldsymbol x}_i,y)\right ]\notin Y_i]\!].
\end{equation}
One-Error evaluates how often the top-ranked predicted label is not included in the set of true labels. A smaller One-Error value indicates better predictive performance, as it implies that the most confident prediction of the model corresponds to a relevant (true) label more frequently. The definition of Ranking Loss is given as follows.
\begin{align}
{rloss}(f)=\frac {1}{p}\sum _{i=1}^p \frac {1}{|Y_i| |\bar Y_i|}|\{(y',y'')\mid f({\boldsymbol x}_i,y')\\ \leq f({\boldsymbol x}_i,y''),~(y',y'')\in Y_i\times \bar Y_i)\}|
\end{align}
Ranking Loss measures the average fraction of label pairs that are incorrectly ordered across all instances. It quantifies how often irrelevant labels are ranked above relevant ones in the prediction output. A smaller Ranking Loss value indicates better performance, implying that the model successfully assigns higher scores to true labels than to false labels in most cases. The definition of Multi-Label Accuracy is given as follows.
\begin{equation}
\text{Multi-label Accuracy}(h) = \frac{1}{p} \sum_{i=1}^{p} 
\frac{|h(\mathbf{x}_i) \cap Y_i|}{|h(\mathbf{x}_i) \cup Y_i|}.
\end{equation}
Multi-Label Accuracy measures the overlap between the predicted and true label sets for each instance and then averages the result over all samples. It is equivalent to the Jaccard similarity coefficient, capturing how similar the prediction and ground truth label sets are. A value of 1 indicates perfect label matching, while 0 represents complete disagreement. Hence, a larger Multi-Label Accuracy value reflects better performance in predicting the correct combination of labels.

Lower values of Hamming Loss, Ranking Loss, and One-Error indicate better performance, whereas higher Multi-label Accuracy implies superior predictive capability of the selected feature subset. Unlike traditional single-label accuracy, these measures collectively reflect both label dependency and prediction consistency, offering a more comprehensive view of feature quality in multi-label environments.

%%%%%%%%%%%%%%%%%%%%%%%%%%%%%%%%%%%%%%%%%%%%%%%%%%%%%%%%%%%%EXPERIMENT%%%%%%%%%%%%%%%%%%%%%%%%%%%%%%%%%%%%%%%%%%%%%%%%%%%%%%%%%%%%
\section{Experiment}
To demonstrate the validity of the proposed evaluation framework, experiments were conducted on 21 publicly available multi-label datasets from diverse domains, including text, biology, image, and signal processing. The datasets were obtained from the Multi-Label Learning Resources repository provided by the University of Córdoba. The names of the datasets are as follows: Inter3000 \cite{inter3000}, CHD49 \cite{chd49}, GpositiveGO \cite{gopositiveplantvrius}, GpositivePseAAC \cite{gopositiveplantvrius}, PlantGO \cite{gopositiveplantvrius}, PlantPseAAC \cite{gopositiveplantvrius}, VirusGO \cite{gopositiveplantvrius}, Waterquality \cite{waterqu}, Birds \cite{birds}, CAL500 \cite{cal500}, Emotions \cite{emotions}, Enron \cite{enron}, Flags \cite{flags}, Foodtruck \cite{foodtruck}, Genbase \cite{genbase}, Image \cite{image}, Langlog \cite{langlog}, Medical \cite{medical}, Scene \cite{scene}, Coffee \cite{coffee}], and Yeast \cite{yeast}. These datasets cover a wide range of label cardinalities and feature dimensions, allowing for a comprehensive comparison across different data characteristics.

To evaluate the effectiveness of the selected feature subsets, Multi-Label k-Nearest Neighbor (MLkNN) \cite{mlknn}, where the number of neighbors was fixed to $k=10$ Each experiment was repeated ten times under a hold-out cross-validation scheme. For each run, 80\% of the instances were randomly selected for training, while the remaining 20\% were used as the test set to assess classification performance.

The predicted labels of the test samples were evaluated using four standard multi-label measures. Among them, Multi-Label Accuracy serves as the primary metric for assessing classification performance, while the complementary loss-based measures, Hamming Loss, Ranking Loss, and One-Error, are provided in Appendix~\ref{appendix} for reference. A higher Multi-Label Accuracy or lower loss-based values indicate better performance, reflecting each model’s ability to capture the structural label dependencies within the datasets.
% ===================== Multi-label accuracy (↑) =====================
\begin{table}[!t]
\setlength{\abovecaptionskip}{0pt}
\setlength{\belowcaptionskip}{0pt}
\centering
\small
\caption{Comparison of EMUFS \citep{pdp} and representative unsupervised FS methods on 21 multi-label datasets using evaluation measure Multi-Label Accuracy. The highest values for accuracy are highlighted in bold. “Avg. Rank” represents the average ranking of each method across all datasets, where a lower value indicates better overall performance.}
\label{tab1}
\resizebox{\textwidth}{!}{%
\begin{tabular}{lcccccc}
\toprule
Datasets & \textbf{EMUFS} & CNAFS & EGCFS & FSDK & MCFS & RUSLP \\
\midrule
 Inter3000   & 0.189$\pm$0.047 &\textbf{0.191$\pm$0.030} & 0.154$\pm$0.034 &0.182$\pm$0.036 &0.167$\pm$0.044 &0.174$\pm$0.042 \\
 CHD49 &0.445$\pm$0.033 & 0.446$\pm$0.052 &0.469$\pm$0.025 & \textbf{0.474$\pm$0.040} &0.441 $\pm$0.047 &0.461$\pm$0.039 \\
 GpositiveGO    &\textbf{0.814$\pm$0.029} &0.350$\pm$0.063 &0.297$\pm$0.046 &0.810$\pm$0.047 &0.808$\pm$0.030 &0.336$\pm$0.062 \\
 GpositivePseAAC    &\textbf{0.618$\pm$0.059} &0.500$\pm$0.071 &0.433$\pm$0.048 &0.621$\pm$0.024 &0.599$\pm$0.045 &0.540$\pm$0.065 \\
 PlantGO    &\textbf{0.658$\pm$0.035} &0.141$\pm$0.019 &0.101$\pm$0.008 &0.584$\pm$0.053 &0.643$\pm$0.041 &0.121$\pm$0.015 \\
 PlantPseAAC &\textbf{0.265$\pm$0.037} &0.139$\pm$0.032 &0.117$\pm$0.020 &0.224$\pm$0.059 & 0.236$\pm$0.039 &0.214$\pm$0.047 \\
 VirusGO    &\textbf{0.680$\pm$0.088} &0.456$\pm$0.086 &0.282$\pm$0.050 &0.634$\pm$0.116 &0.709$\pm$0.067 &0.456$\pm$0.090 \\
 Water quality &0.401$\pm$0.014 &0.397$\pm$0.017 &\textbf{0.402$\pm$0.017} &0.399$\pm$0.014 &0.395$\pm$0.004 &0.406$\pm$0.012 \\
 Birds &0.371$\pm$0.145 &0.365$\pm$0.091 &0.265$\pm$0.091 &0.459$\pm$0.121 & \textbf{0.485$\pm$0.032} &0.288$\pm$0.082 \\
 CAL500 &0.173$\pm$0.004 &0.179$\pm$0.007 &0.173$\pm$0.004 &\textbf{0.185$\pm$0.005} &0.179$\pm$0.006 &0.179$\pm$0.008 \\
 Emotions & 0.508$\pm$0.034 &\textbf{0.540$\pm$0.024} &0.528$\pm$0.022 &0.530$\pm$0.031 &0.521$\pm$0.014 &0.53$\pm$0.031 \\
 Enron &\textbf{0.269$\pm$0.027} &0.175$\pm$0.023 &0.203$\pm$0.029 &0.258$\pm$0.032 &0.182$\pm$0.013 &0.087$\pm$0.003 \\
 Flags &\textbf{0.525$\pm$0.040} &0.513$\pm$0.023 &0.520$\pm$0.031 &0.506$\pm$0.034 &0.499$\pm$0.041 &0.514$\pm$0.034 \\
 Foodtruck &0.254$\pm$0.023 &0.245$\pm$0.023 &\textbf{0.275$\pm$0.022} &0.246$\pm$0.020 &0.267$\pm$0.034 &0.254$\pm$0.023 \\
 Genbase &0.318$\pm$0.039 &0.199$\pm$0.059 &\textbf{0.358$\pm$0.091} &0.274$\pm$0.093 &0.221$\pm$0.065 &0.375$\pm$0.208 \\
 Image &0.495$\pm$0.030 &\textbf{0.544$\pm$0.025} &0.505$\pm$0.011 &0.475$\pm$0.032 &0.527$\pm$0.022 &0.505$\pm$0.012 \\
 Llog &\textbf{0.042$\pm$0.005} &0.022$\pm$0.006 &0.024$\pm$0.005 &0.064$\pm$0.058 &0.050$\pm$0.048 &0.017$\pm$0.001 \\
 Medical &\textbf{0.433$\pm$0.083} &0.032$\pm$0.003 &0.033$\pm$0.002 &0.373$\pm$0.129 &0.404$\pm$0.142 &0.031$\pm$0.001 \\
 Scene &0.518$\pm$0.027 &0.645$\pm$0.020 &0.559$\pm$0.016 &0.545$\pm$0.025 &\textbf{0.668$\pm$0.022} &0.576$\pm$0.017 \\
 Coffee &\textbf{0.067$\pm$0.013} &0.027$\pm$0.007 &0.026$\pm$0.003 &0.059$\pm$0.013 &0.065$\pm$0.015 &0.018$\pm$0.003 \\
 Yeast & 0.412$\pm$0.013 &0.410$\pm$0.029 &0.374$\pm$0.016 &0.399$\pm$0.035 &0.427$\pm$0.031 &\textbf{0.435$\pm$0.028} \\
\midrule
 Avg.Rank & \textbf{2.76} & 3.86 & 4.24 & 3.05 & 3.05 & 3.67 \\
\bottomrule
\end{tabular}%
}
\end{table}

The experimental results summarized in Table \ref{tab1} demonstrate the comparative performance of the Entropy Maximization UFS (EMUFS) \cite{pdp} and several representative unsupervised FS methods across multi-label datasets. Overall, the EMUFS achieves competitive or superior performance, ranking first in Multi-Label Accuracy followed by MCFS \cite{mcfs}, Fast Sparse Discriminative K-means (FSDK) \cite{fsdk}, Robust Unsupervised Feature Selection with Local Preserving (RUSLP) \cite{ruslp}, Convex Nonnegative Matrix Factorization with Adaptive Graph Constraint (CNAFS) \cite{cnafs}, and Novel Unsupervised Feature Selection via Adaptive Graph Learning and Constraint (EGCFS) \cite{egcfs} and showing the lowest average ranks in Hamming Loss, Ranking Loss, and One-Error EMUFS, MCFS, FSDK, CNAFS, RUSLP, EGCFS. These results highlight the robustness of the EMUFS across diverse domains. While FSDK and MCFS sometimes exhibit coPrevious studies have commonly reported that recently developed methods, such as FSDK, RUSLP, and CNAFS, outperform traditional graph-based methods like MCFS in single-label environments.amming Loss, Ranking Loss, and One-Error) are provided in Appendix~\ref{appendix}, whereas Table \ref{tab1} focuses on Multi-Label Accuracy as the primary indicator of overall effectiveness.

Previous studies have commonly reported that recently developed methods such as FSDK, RUSLP, and CNAFS outperform traditional graph-based methods like MCFS in single-label environments. These findings were largely derived under evaluation frameworks that assume a one-to-one correspondence between features and class variables, favoring information-theoretic or sparse-representation–based approaches that optimize discriminability for individual labels. However, when reexamined under a multi-label evaluation setting, this relative superiority no longer holds consistently. In particular, MCFS, despite being a comparatively classical method, demonstrates competitive or even superior performance across multiple evaluation criteria. This observation suggests that single-label evaluations may have overestimated the generalization capability of newer methods by failing to reflect inter-label dependencies inherent in real-world data. Therefore, the conventional assumption may be an artifact of the single-label paradigm, emphasizing the need to reassess FS methods under a unified multi-label perspective. These results confirm that performance rankings reported in traditional single-label evaluations do not necessarily hold under multi-label conditions.
%%%%%%%%%%%%%%%%%%%%%%%%%%%%%%%%%%%%%%%%%%%%%%%%%%%%%%%%%%%%CONCLUSION%%%%%%%%%%%%%%%%%%%%%%%%%%%%%%%%%%%%%%%%%%%%%%%%%%%%%%%%%%%%
\section{Conclusion}
This study revisited the evaluation paradigm of UFS by examining existing methods under a multi-label classification framework. While previous UFS research predominantly relied on single-label evaluations, our findings on 21 diverse datasets revealed that feature subsets produce substantially different results when assessed under multi-label conditions. This discrepancy indicates that the reported superiority of UFS methods in earlier studies may be influenced by random label assignments rather than their true structural representation capability. The results emphasize the necessity of employing multi-label evaluation protocols to more accurately reflect the generalization and robustness of FS methods in real-world data scenarios.

\begin{ack}
This research was supported by the Institute of Information \& Communications Technology Planning \& Evaluation (IITP) grant funded by the Korea government (MSIT) [RS-2021-II211341, Artificial Intelligent Graduate School Program (Chung-Ang University)].
\end{ack}

%%%%%%%%%%%%%%%%%%%%%%%%%%%%%%%%%%%%%%%%%%%%%%%%%%%%%%%%%%%%APPENDIX%%%%%%%%%%%%%%%%%%%%%%%%%%%%%%%%%%%%%%%%%%%%%%%%%%%%%%%%%%%%
\appendix
\section{Additional Evaluation Measures}
\label{appendix}
% ===================== Hamming loss (↓) =====================
\begin{table}[H]
\centering
\small
\caption{Comparison of EMUFS \citep{pdp} and representative unsupervised FS methods on 21 multi-label datasets using evaluation measure Hamming Loss. The lowest values for accuracy are highlighted in bold. “Avg. Rank” represents the average ranking of each method across all datasets, where a lower value indicates better overall performance.}
\label{tab2}
\resizebox{\textwidth}{!}{%
\begin{tabular}{lcccccc}
\toprule
Datasets & \textbf{EMUFS} & CNAFS & EGCFS & FSDK & MCFS & RUSLP \\
\midrule
Inter3000 & \textbf{0.383$\pm$0.037} &0.414$\pm$0.062 &0.431$\pm$0.042 &0.388$\pm$0.041 &0.411$\pm$0.053 &0.448$\pm$0.046 \\
CHD49 &0.397$\pm$0.055 &0.404$\pm$0.068 &0.367$\pm$0.021 &\textbf{0.359$\pm$0.050} &0.404$\pm$0.071 &0.368$\pm$0.040 \\
GpositiveGO & 0.095$\pm$0.015 &0.485$\pm$0.120 &0.586$\pm$0.114 &\textbf{0.092$\pm$0.022} &0.099$\pm$0.021 &0.492$\pm$0.152 \\
GpositivePseAAC &\textbf{0.189$\pm$0.031} &0.261$\pm$0.042 &0.304$\pm$0.030 &0.190$\pm$0.016 &0.207$\pm$0.028 &0.237$\pm$0.039 \\
PlantGO &\textbf{0.059$\pm$0.007} & 0.443$\pm$0.090 &0.762$\pm$0.085 &0.077$\pm$0.012 &0.067$\pm$0.007 &0.569$\pm$0.089 \\
PlantPseAAC &\textbf{0.161$\pm$0.023} &0.233$\pm$0.036 &0.287$\pm$0.037 &0.189$\pm$0.048 &0.171$\pm$0.034 & 0.183$\pm$0.035 \\
VirusGO &0.121$\pm$0.038 &0.228$\pm$0.059 &0.456$\pm$0.085 &0.138$\pm$0.039 &\textbf{0.111$\pm$0.025} &0.225$\pm$0.050 \\
Water quality &0.336$\pm$0.010 & 0.336$\pm$0.007 &0.336$\pm$0.013 &0.336$\pm$0.010 &0.336$\pm$0.007 &0\textbf{.333$\pm$0.010} \\
Birds &0.104$\pm$0.021 &0.104$\pm$0.008 &0.112$\pm$0.017 &0.092$\pm$0.016 &\textbf{0.085$\pm$0.009} &0.109$\pm$0.013 \\
CAL500 &0.336$\pm$0.011 &0.324$\pm$0.012 &0.333$\pm$0.007 &0.316$\pm$0.014 &\textbf{0.323$\pm$0.012} &0.325$\pm$0.017 \\
Emotions &0.242$\pm$0.021 & \textbf{0.231$\pm$0.014} & 0.241$\pm$0.012 &0.238$\pm$0.016 &0.244$\pm$0.016 &0.235$\pm$0.018 \\
Enron &\textbf{0.109$\pm$0.006} &0.179$\pm$0.027 &0.133$\pm$0.018 &0.112$\pm$0.010 &0.160$\pm$0.020 &0.594$\pm$0.037 \\
Flags &\textbf{0.340$\pm$0.036} &0.345$\pm$0.022 &0.346$\pm$0.021 & 0.348$\pm$0.032 & 0.354$\pm$0.036 &0.348$\pm$0.025 \\
Foodtruck &0.331$\pm$0.038 &0.333$\pm$0.022 &\textbf{0.301$\pm$0.029} &0.329$\pm$0.023 &0.310$\pm$0.044 &0.340$\pm$0.022 \\
Genbase & 0.093$\pm$0.018 &0.189$\pm$0.060 &\textbf{0.083$\pm$0.021} &0.127$\pm$0.039 &0.169$\pm$0.059 &0.093$\pm$0.042 \\
Image &0.232$\pm$0.015 &\textbf{0.212$\pm$0.012} &0.243$\pm$0.008 &0.247$\pm$0.018 &0.223$\pm$0.015 &0.240$\pm$0.007 \\
Llog &\textbf{0.089$\pm$0.010} &0.170$\pm$0.048 &0.121$\pm$0.024 &0.098$\pm$0.025 &0.091$\pm$0.010 &0.851$\pm$0.023 \\
Medical &\textbf{0.036$\pm$0.010} &0.743$\pm$0.078 &0.682$\pm$0.033 &0.047$\pm$0.021 &0.045$\pm$0.019 &0.789$\pm$0.046 \\
Scene & 0.177$\pm$0.011 &0.120$\pm$0.007 &0.152$\pm$0.007 &0.159$\pm$0.008 &\textbf{0.112$\pm$0.008} &0.147$\pm$0.007 \\
Coffee &\textbf{0.083$\pm$0.007} &0.173$\pm$0.035 &0.181$\pm$0.034 &0.087$\pm$0.012 &0.086$\pm$0.007 &0.341$\pm$0.054 \\
Yeast &0.306$\pm$0.010 & 0.312$\pm$0.033 &0.350$\pm$0.023 &0.318$\pm$0.033 &0.294$\pm$0.034 &\textbf{0.288$\pm$0.025} \\
\midrule
Avg.Rank & \textbf{2.43} & 3.81 & 4.29 & 2.95 & 2.76 & 4.10 \\
\bottomrule
\end{tabular}
}
\end{table}
Appendix~\ref{appendix} presents the supplementary experimental results evaluated using loss-based multi-label measures, including Hamming Loss, Ranking Loss, and One-Error. These measures provide complementary perspectives to the main analysis, quantifying classification consistency and label-ranking stability. While the primary paper focuses on Multi-Label Accuracy as the most interpretable and representative indicator, the trends across these additional metrics are largely consistent with the overall findings, further validating the robustness of the proposed evaluation.
% ===================== Ranking loss (↓) =====================
\begin{table}[!t]
\centering
\small
\caption{Comparison of EMUFS \citep{pdp} and representative unsupervised FS methods on 21 multi-label datasets using evaluation measure Ranking Loss. The lowest values for accuracy are highlighted in bold. “Avg. Rank” represents the average ranking of each method across all datasets, where a lower value indicates better overall performance.}
\label{tab3}
\resizebox{\textwidth}{!}{%
\begin{tabular}{lcccccc}
\toprule
Datasets & \textbf{EMUFS} & CNAFS & EGCFS & FSDK & MCFS & RUSLP \\
\midrule
Inter3000 & \textbf{0.418$\pm$0.045} & 0.438$\pm$0.051 & 0.432$\pm$0.048 & 0.437$\pm$0.048 & 0.455$\pm$0.058 & 0.455$\pm$0.055 \\
CHD49 & 0.240$\pm$0.015 & 0.228$\pm$0.027 & \textbf{0.222$\pm$0.018} & 0.231$\pm$0.021 & 0.235$\pm$0.015 & 0.232$\pm$0.023 \\
GpositiveGO & 0.075$\pm$0.018 & 0.270$\pm$0.062 & 0.279$\pm$0.036 & 0.074$\pm$0.020 & \textbf{0.069$\pm$0.010} & 0.237$\pm$0.058 \\
GpositivePseAAC & 0.145$\pm$0.023 & 0.222$\pm$0.052 & 0.239$\pm$0.023 & \textbf{0.145$\pm$0.018} & 0.158$\pm$0.028 & 0.204$\pm$0.053 \\
PlantGO & \textbf{0.049$\pm$0.008} & 0.253$\pm$0.018 & 0.271$\pm$0.017 & 0.075$\pm$0.017 & 0.070$\pm$0.012 & 0.259$\pm$0.020 \\
PlantPseAAC & \textbf{0.188$\pm$0.007} & 0.234$\pm$0.025 & 0.244$\pm$0.018 & 0.192$\pm$0.017 & 0.196$\pm$0.022 & 0.217$\pm$0.019 \\
VirusGO & \textbf{0.076$\pm$0.032} & 0.142$\pm$0.033 & 0.229$\pm$0.047 & 0.094$\pm$0.041 & 0.091$\pm$0.017 & 0.132$\pm$0.025 \\
Waterquality & 0.260$\pm$0.012 & 0.263$\pm$0.010 & 0.259$\pm$0.009 & 0.261$\pm$0.012 & 0.262$\pm$0.010 & \textbf{0.253$\pm$0.006} \\
Birds & 0.205$\pm$0.035 & 0.203$\pm$0.025 & \textbf{0.191$\pm$0.017} & 0.208$\pm$0.027 & 0.197$\pm$0.027 & 0.193$\pm$0.019 \\
Cal500 & 0.188$\pm$0.004 & 0.185$\pm$0.003 & 0.185$\pm$0.005 & 0.184$\pm$0.007 & \textbf{0.182$\pm$0.003} & 0.185$\pm$0.005 \\
Emotions & \textbf{0.166$\pm$0.017} & 0.173$\pm$0.016 & 0.177$\pm$0.022 & 0.174$\pm$0.020 & 0.188$\pm$0.015 & 0.173$\pm$0.016 \\
Enron & \textbf{0.095$\pm$0.003} & 0.109$\pm$0.006 & 0.104$\pm$0.004 & 0.098$\pm$0.004 & 0.107$\pm$0.007 & 0.111$\pm$0.005 \\
Flags & 0.232$\pm$0.029 & \textbf{0.230$\pm$0.022} & 0.236$\pm$0.017 & 0.247$\pm$0.020 & 0.251$\pm$0.022 & 0.233$\pm$0.033 \\
Foodtruck & 0.170$\pm$0.022 & 0.169$\pm$0.035 & 0.170$\pm$0.020 & 0.168$\pm$0.013 & \textbf{0.161$\pm$0.019} & 0.173$\pm$0.026 \\
Genbase & 0.009$\pm$0.002 & 0.045$\pm$0.019 & 0.009$\pm$0.006 & 0.013$\pm$0.007 & 0.012$\pm$0.007 & \textbf{0.006$\pm$0.005} \\
Image & 0.213$\pm$0.021 & \textbf{0.189$\pm$0.014} & 0.217$\pm$0.010 & 0.231$\pm$0.018 & 0.197$\pm$0.011 & 0.228$\pm$0.012 \\
Langlog & 0.179$\pm$0.010 & 0.179$\pm$0.010 & 0.187$\pm$0.017 & \textbf{0.178$\pm$0.011} & 0.183$\pm$0.011 & 0.187$\pm$0.012 \\
Medical & 0.052$\pm$0.009 & 0.133$\pm$0.005 & 0.119$\pm$0.007 & 0.054$\pm$0.008 & \textbf{0.042$\pm$0.004} & 0.135$\pm$0.008 \\
Scene & 0.186$\pm$0.013 & 0.102$\pm$0.009 & 0.140$\pm$0.008 & 0.145$\pm$0.014 & \textbf{0.094$\pm$0.006} & 0.129$\pm$0.010 \\
Stackexcoffee & 0.295$\pm$0.034 & 0.273$\pm$0.045 & 0.271$\pm$0.045 & 0.257$\pm$0.027 & \textbf{0.252$\pm$0.031} & 0.305$\pm$0.047 \\
Yeast & 0.190$\pm$0.005 & 0.179$\pm$0.005 & 0.193$\pm$0.006 & 0.182$\pm$0.013 & 0.174$\pm$0.009 & \textbf{0.171$\pm$0.005} \\
\midrule
Avg.Rank & \textbf{2.90} & 3.67 & 3.95 & 3.24 & 2.95 & 3.81 \\
\bottomrule
\end{tabular}
}
\end{table}

% ===================== One error (↓) =====================
\begin{table}[!t]
\centering
\small
\caption{Comparison of EMUFS \citep{pdp} and representative unsupervised FS methods on 21 multi-label datasets using evaluation measure One-Error. The lowest values for accuracy are highlighted in bold. “Avg. Rank” represents the average ranking of each method across all datasets, where a lower value indicates better overall performance.}
\label{tab4}
\resizebox{\textwidth}{!}{%
\begin{tabular}{lcccccc}
\toprule
Datasets & \textbf{EMUFS} & CNAFS & EGCFS & FSDK & MCFS & RUSLP \\
\midrule
Inter3000 & \textbf{0.742$\pm$0.066} & 0.758$\pm$0.052 & 0.767$\pm$0.074 & 0.752$\pm$0.047 & 0.788$\pm$0.062 & 0.797$\pm$0.064 \\
CHD49 & 0.273$\pm$0.035 & 0.261$\pm$0.051 & 0.269$\pm$0.042 & 0.262$\pm$0.029 & 0.260$\pm$0.043 & \textbf{0.258$\pm$0.037} \\
GpositiveGO & 0.171$\pm$0.033 & 0.549$\pm$0.111 & 0.528$\pm$0.075 & 0.164$\pm$0.037 & \textbf{0.134$\pm$0.018} & 0.477$\pm$0.100 \\
GpositivePseAAC & \textbf{0.287$\pm$0.039} & 0.430$\pm$0.088 & 0.465$\pm$0.038 & 0.291$\pm$0.035 & 0.297$\pm$0.051 & 0.397$\pm$0.083 \\
PlantGO & \textbf{0.250$\pm$0.020} & 0.821$\pm$0.026 & 0.896$\pm$0.015 & 0.318$\pm$0.042 & 0.301$\pm$0.034 & 0.865$\pm$0.083 \\
PlantPseAAC & \textbf{0.629$\pm$0.037} & 0.707$\pm$0.048 & 0.711$\pm$0.030 & 0.631$\pm$0.036 & 0.635$\pm$0.022 & 0.673$\pm$0.026 \\
VirusGO & \textbf{0.185$\pm$0.069} & 0.420$\pm$0.073 & 0.561$\pm$0.102 & 0.244$\pm$0.098 & 0.222$\pm$0.035 & 0.407$\pm$0.100 \\
Waterquality & 0.312$\pm$0.030 & 0.320$\pm$0.032 & 0.298$\pm$0.025 & 0.307$\pm$0.036 & 0.312$\pm$0.025 & \textbf{0.290$\pm$0.034} \\
Birds & 0.571$\pm$0.057 & 0.568$\pm$0.064 & 0.578$\pm$0.034 & 0.514$\pm$0.036 & \textbf{0.488$\pm$0.042} & 0.534$\pm$0.063 \\
Cal500 & 0.190$\pm$0.037 & 0.195$\pm$0.038 & 0.186$\pm$0.042 & 0.190$\pm$0.041 & \textbf{0.179$\pm$0.052} & 0.194$\pm$0.064 \\
Emotions & 0.304$\pm$0.029 & \textbf{0.284$\pm$0.033} & 0.298$\pm$0.048 & 0.291$\pm$0.048 & 0.312$\pm$0.036 & 0.308$\pm$0.041 \\
Enron & \textbf{0.300$\pm$0.025} & 0.423$\pm$0.042 & 0.329$\pm$0.027 & 0.337$\pm$0.032 & 0.378$\pm$0.048 & 0.442$\pm$0.046 \\
Flags & \textbf{0.224$\pm$0.076} & 0.239$\pm$0.079 & 0.268$\pm$0.074 & 0.271$\pm$0.087 & 0.263$\pm$0.025 & 0.239$\pm$0.050 \\
Foodtruck & 0.288$\pm$0.047 & 0.283$\pm$0.059 & 0.290$\pm$0.062 & 0.293$\pm$0.049 & \textbf{0.281$\pm$0.045} & 0.294$\pm$0.047 \\
Genbase & 0.021$\pm$0.013 & 0.355$\pm$0.131 & \textbf{0.014$\pm$0.015} & 0.022$\pm$0.008 & 0.024$\pm$0.018 & 0.021$\pm$0.022 \\
Image & 0.398$\pm$0.037 & \textbf{0.352$\pm$0.030} & 0.394$\pm$0.015 & 0.424$\pm$0.031 & 0.362$\pm$0.014 & 0.411$\pm$0.014 \\
Langlog & \textbf{0.837$\pm$0.022} & 0.866$\pm$0.026 & 0.885$\pm$0.026 & 0.858$\pm$0.027 & 0.860$\pm$0.022 & 0.876$\pm$0.038 \\
Medical & 0.336$\pm$0.022 & 0.717$\pm$0.029 & 0.701$\pm$0.020 & 0.320$\pm$0.027 & \textbf{0.242$\pm$0.019} & 0.718$\pm$0.012 \\
Scene & 0.432$\pm$0.029 & 0.308$\pm$0.021 & 0.386$\pm$0.020 & 0.402$\pm$0.034 & \textbf{0.280$\pm$0.020} & 0.361$\pm$0.021 \\
Stackexcoffee & \textbf{0.698$\pm$0.049} & 0.793$\pm$0.057 & 0.782$\pm$0.043 & 0.760$\pm$0.049 & 0.704$\pm$0.080 & 0.816$\pm$0.049 \\
Yeast & \textbf{0.259$\pm$0.015} & 0.263$\pm$0.020 & 0.277$\pm$0.021 & 0.263$\pm$0.018 & 0.272$\pm$0.013 & 0.264$\pm$0.024 \\
\midrule
Avg.Rank & \textbf{2.52} & 3.90 & 4.29 & 3.14 & 2.76 & 4.14 \\
\bottomrule
\end{tabular}
}
\end{table}

\end{document}